\documentclass[a4paper,conference]{IEEEtran}
\usepackage{cite}
\usepackage{graphicx}
\usepackage{amsmath}
\usepackage{algorithmic}
\usepackage{array}
\usepackage{subfig}
\usepackage{url}

\begin{document}

\title{Visual Representations of Physiological Signals for Fake Video Detection}

\author{\IEEEauthorblockN{Kalin Stefanov}
\IEEEauthorblockA{Monash University\\
Australia\\
kalin.stefanov@monash.edu}
\and
\IEEEauthorblockN{Bhawna Paliwal}
\IEEEauthorblockA{IIT Ropar\\
India\\
2017chb1039@iitrpr.ac.in}
\and
\IEEEauthorblockN{Abhinav Dhall}
\IEEEauthorblockA{IIT Ropar\\
India\\
abhinav@iitrpr.ac.in}}

\maketitle

\newcommand{\etal}{\textit{et al}.~}
\newcommand{\ie}{\textit{i}.\textit{e}.,~}
\newcommand{\eg}{\textit{e}.\textit{g}.,~}

\begin{abstract}
Realistic fake videos are a potential tool for spreading harmful misinformation given our increasing online presence and information intake. This paper presents a multimodal learning-based method for detection of real and fake videos. The method combines information from three modalities -- audio, video, and physiology. We investigate two strategies for combining the video and physiology modalities, either by augmenting the video with information from the physiology or by novelly learning the fusion of those two modalities with a proposed Graph Convolutional Network architecture. Both strategies for combining the two modalities rely on a novel method for generation of visual representations of physiological signals. The detection of real and fake videos is then based on the dissimilarity between the audio and modified video modalities. The proposed method is evaluated on two benchmark datasets and the results show significant increase in detection performance compared to previous methods.
\end{abstract}

\section{Introduction}
\label{sec:introduction}
Advances in computer vision and deep learning have enabled the creation of very realistic fake versions of videos, known as \textit{deepfakes}\footnote{https://github.com/deepfakes/faceswap}\footnote{https://github.com/dfaker/df}\footnote{https://github.com/iperov/DeepFaceLab}\footnote{https://github.com/shaoanlu/faceswap-GAN}. Highly realistic deepfakes are a potential tool for spreading harmful misinformation given our increasing online presence and information intake. Hence, it has become increasingly important to identify deepfakes with more accurate and reliable methods. The recent surge in synthesized fake video content on the Internet has also led to the release of several benchmark datasets (\eg~\cite{Dolhansky2020DFDC,Korshunov2018DeepFakes,Roessler2019FaceForensics}) and methods (\eg~\cite{Roessler2019FaceForensics,Verdoliva2019Multimedia,Yang2019Exposing}) for fake content detection. The aim of these fake video detection methods is to correctly classify any given input video as either real or fake.

There is rich and growing literature on different methods for fake video detection. Previous methods can be loosely placed in two groups -- methods that use a single modality (unimodal) and methods that use more than one modality (multimodal) for the detection of fake videos. Unimodal methods usually exploit artifacts in the visual modality, whereas, multimodal methods combine multiple modalities (\eg audio and video) that can provide complementary information and lead to better fake video detection rates. 

The main idea of this work is to validate the hypothesis that \textit{physiological signals} information from face videos can be used to better classify fake and real videos, and to answer the related questions such as how can we \textit{generate} suitable representations of these physiological signals, and how can we \textit{incorporate} these signals in existing methods to improve the performance even further. The idea that fake videos exhibit considerably more inconsistencies in the associated physiological signals compared to real videos is thoroughly investigated in this work through evaluation of a video augmentation approach with novel visual representations of physiological signals and a proposed learning-based method. The main contributions of this work include:

\begin{itemize}
\item{A method for generation of visual representations of physiological signals (\ie physiological maps) which are helpful in better determining if a video is real or fake.}
\item{A data augmentation strategy that employs the physiological maps as facial ``heat maps''.}
\item{A Graph Convolutional Network for learning-based fusion of the physiological maps and face videos instead of direct augmentation.}
\item{The developed multimodal methods are evaluated on two benchmark deepfake detection datasets and the results show that the proposed physiological maps increase the performance of fake video detection using both strategies; with Graph Convolutional Network based learning giving improvements over the augmentation strategy.}
\end{itemize}

\begin{figure*}[ht]
\centering
\includegraphics[width=\textwidth]{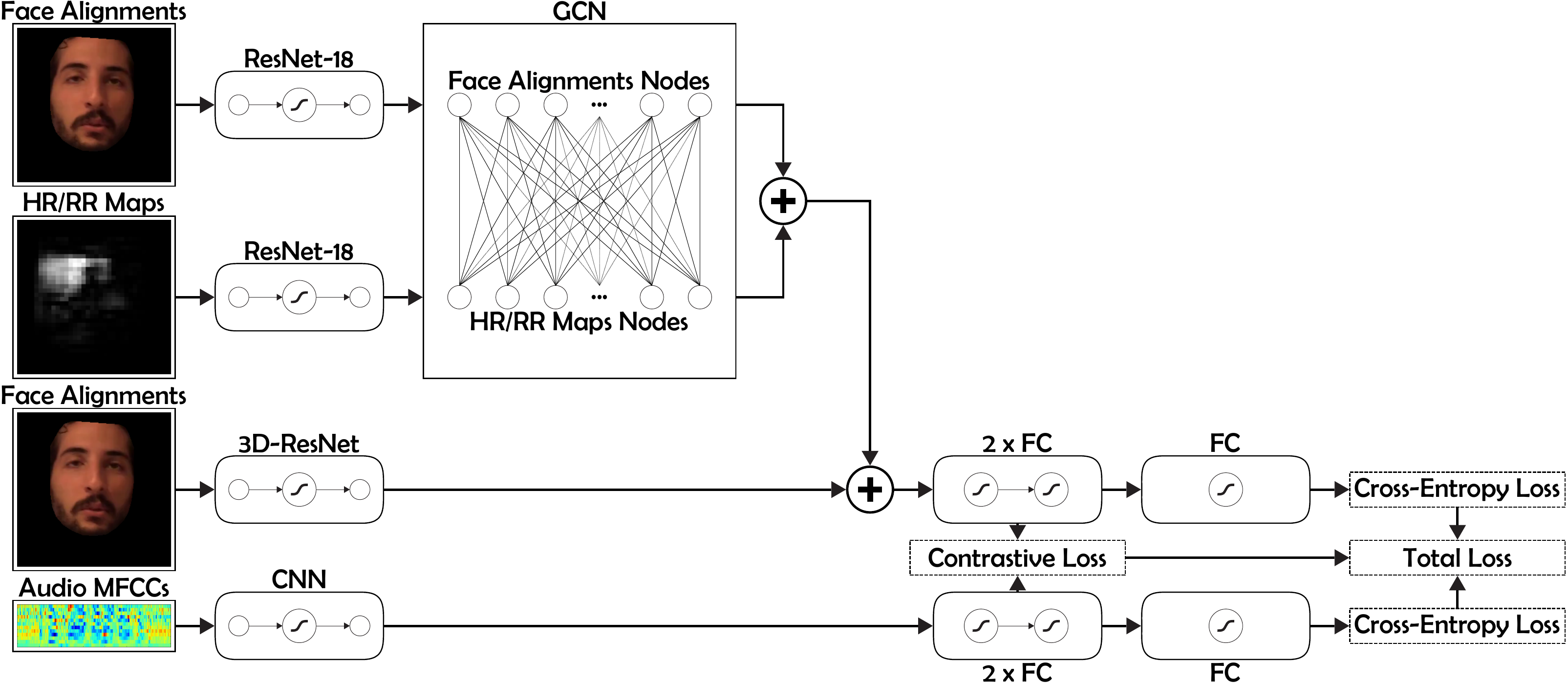}
\caption{Overview of the proposed method for physiology-based fake video detection. Graph Convolution Network is used to learn the fusion of the physiological maps and video streams. Then a contrastive loss based learning is performed incorporating information from the audio stream. Symbol {$\bigoplus$} indicates concatenation of vectors.}
\label{fig:graph_conv_arch}
\end{figure*}

\section{Background}
\label{sec:background}
The topic of deepfake detection is well-researched and growing with increasing use cases~\cite{Zheng2019Survey,Mirsky2021Creation}. This section reviews some of the detection methods that specifically deal with deepfake content of human faces. Deepfake generation methods often generate artifacts in the resulting video that are too subtle for humans but can be easily detected using machine learning. According to~\cite{Mirsky2021Creation} there are two groups of visual artifacts: 1) spatial artifacts and 2) temporal artifacts.

Blending is a spatial artifact related to the areas where the generated content is blended back into the original content. Researchers have proposed different strategies to emphasize these artifacts, for example, edge detectors (\eg~\cite{Mo2018Fake}) and frequency analysis (\eg~\cite{Durall2020Unmasking}). Environment is another spatial artifact related to the content of the fake face being anomalous in the context of the rest of the image. Residuals from the face warping processes (\eg~\cite{Li2019Exposing}) and lighting (\eg~\cite{Straub2019Using}) can indicate the presence of generated content in the image.

Behavior can exhibit temporal artifacts in fake videos. Given sufficient amount of data for a target person, different behaviors can be monitored for anomalies (\eg~\cite{Agarwal2019Protecting}). Another temporal artifacts are related to physiology, for example, blood volume patterns under the skin~\cite{Ciftci2020FakeCatcher}. Synchronization inconsistencies (between audio and video) are another temporal artifact usually found in fake videos (\eg~\cite{Korshunov2018Speaker}). Realistic temporal coherence is challenging to generate, and some previous work focus on the resulting artifacts to detect the fake content (\eg~\cite{Guera2018Deepfake}).

Another approach to fake video detection is to train a generic classifier and let the model choose the features to analyze instead of focusing on the specific artifacts generated by deepfake methods. Deep neural networks have been shown to effectively detect deepfake videos when employed as classifiers (\eg~\cite{Afchar2018MesoNet,Ding2019Swapped,Tariq2018Detecting}). In contrast to classification, anomaly detection methods are trained on real data and then detect outliers during testing (\eg~\cite{Wang2020FakeSpotter,Khalid2020FakeDect}).

The work described in this paper falls under the group of temporal artifacts. More specifically, the hypothesis as earlier mentioned is that fake videos exhibit considerably more temporal inconsistencies in the associated physiological signals compared to real videos. The literature on fake video detection using physiological signals is quite limited compared to other type of spatial and temporal artifacts. The closest related work is the one in~\cite{Ciftci2020FakeCatcher}. The differences between this work and~\cite{Ciftci2020FakeCatcher} include:

\begin{itemize}
\item{This work does not consider specific face regions for extraction of the physiological signals, hence it is not limited to only portrait videos as in~\cite{Ciftci2020FakeCatcher}.}
\item{This work proposes method for generation of physiological maps that can be directly applied as ``heat maps'' on the face videos, producing novel data augmentations.}
\item{This work proposes a multimodal method whereas the method in~\cite{Ciftci2020FakeCatcher} is vision-only.}
\end{itemize}

\begin{figure*}[ht]
\captionsetup[subfigure]{justification=centering}
\centering
\subfloat[Reference responses. \label{subfig:phys_maps_reference}]{\includegraphics[scale=0.485]{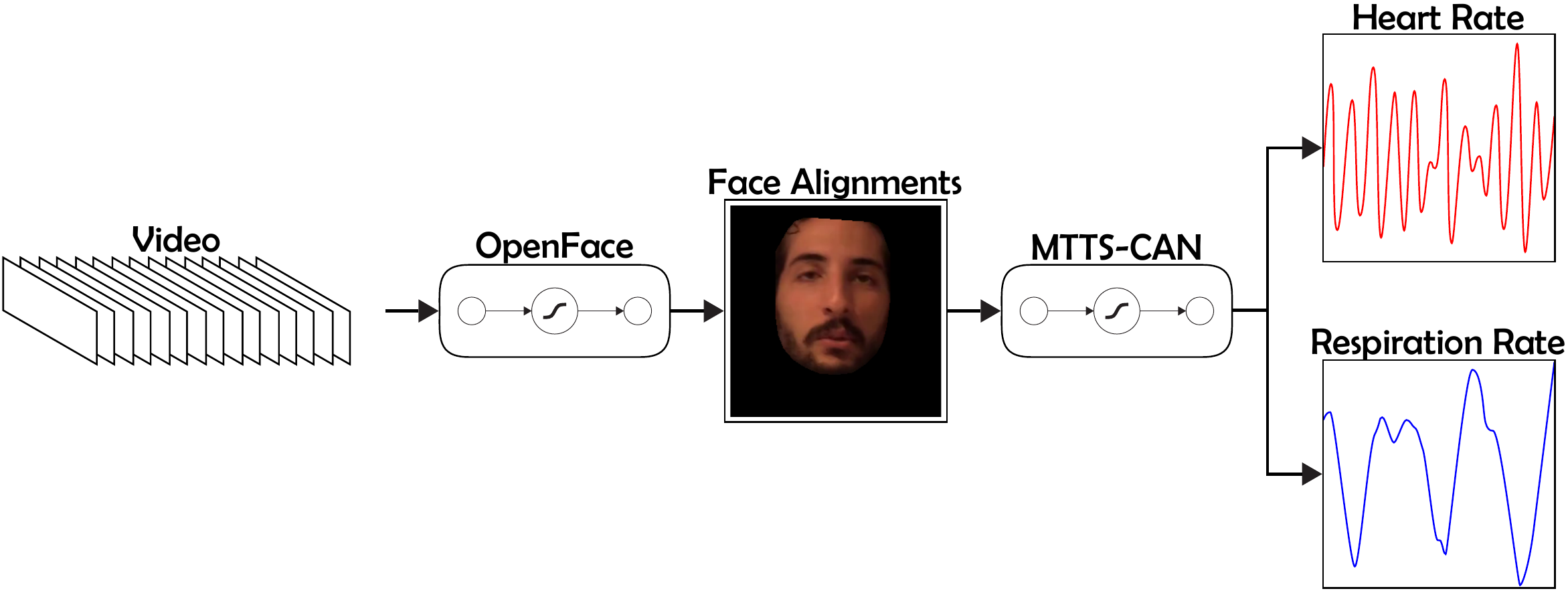}}
\hfill
\subfloat[Difference responses. \label{subfig:phys_maps_difference}]{\includegraphics[scale=0.485]{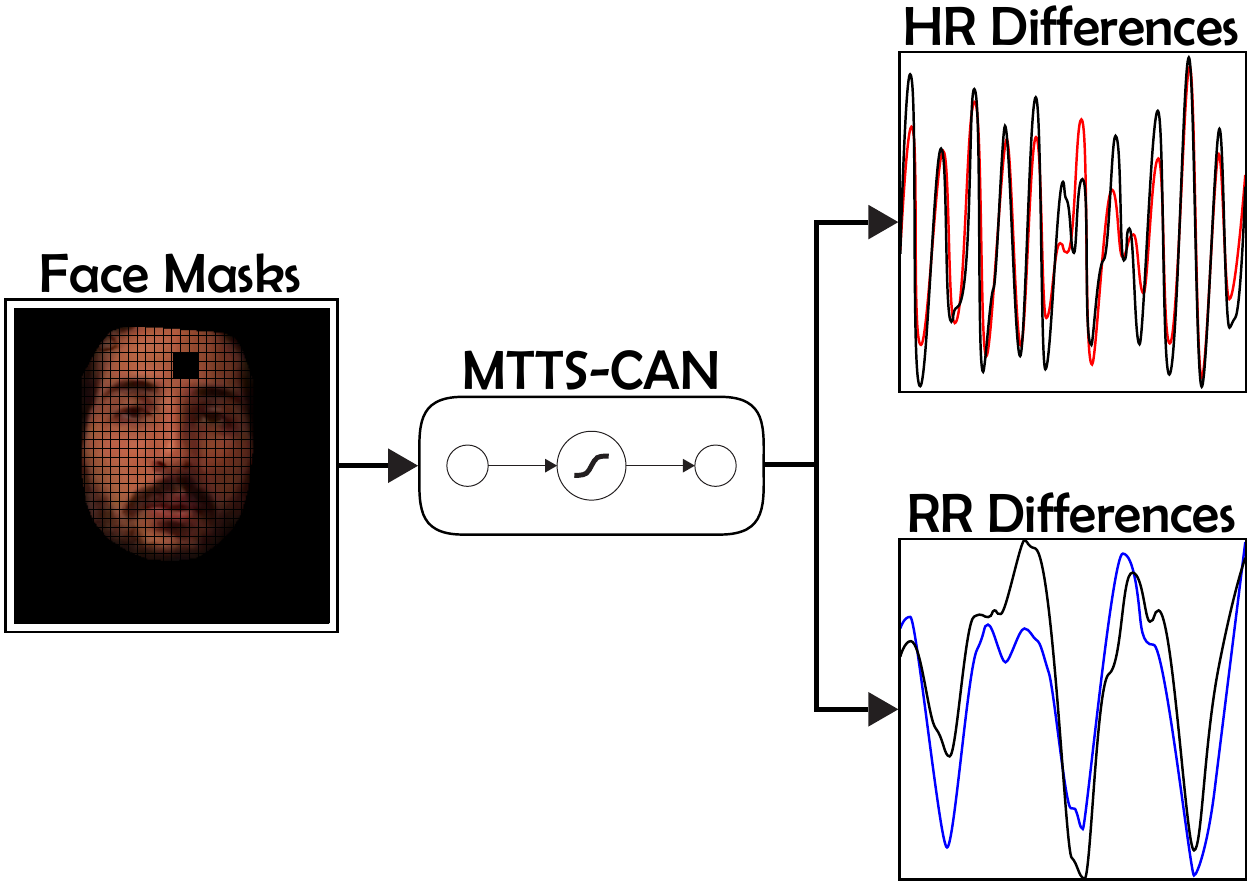}}
\caption{Overview of the proposed method for generation of physiological maps. The aligned face crops are passed through the camera-based physiological measurement model (\ie MTTS-CAN~\cite{Liu2020Multi}) to obtain the reference responses. Then an occlusion patch is slid over the aligned face crops, new responses are calculated and the differences with the reference responses are used as encoding for the relative importance of the image area under the occlusion patch.}
\label{fig:phys_maps_gen}
\end{figure*}

\section{Physiology-Based Fake Video Detection}
\label{sec:physiology-based_fake_video_detection}
The proposed fake video detection method is based on the hypothesis that fake videos exhibit considerably more inconsistencies in the associated camera-based physiological measurements compared to real videos. This section offers a description of the proposed method for physiology-based fake video detection.

\subsection{Method Overview}
Given an input video, the goal is to classify it as real or fake, that is, given a training dataset $D^{train}=\{(v^{1},y^{1}), (v^{2},y^{2}),...,(v^{N},y^{N})\}$ consisting of $N$ videos, where $v^{i}$ denotes the input video and the label $y^{i} \in \{0,1\}$ indicates whether the video is real ($y^{i}=0$) or fake ($y^{i}=1$), the goal is to create a computational model that can correctly classify a new video as either real or fake.

The process of data preparation and classifier training is similar to the work described in~\cite{Chugh2020Not}. The audio signal is extracted from the input video $v^{i}$ using the FFmpeg\footnote{http://ffmpeg.org} library, and then split into $D$-second long segments. Similarly, the video signal is divided into $D$-second long segments, and face tracking is performed on those video segments using the S3FD~\cite{Zhang2017S3FD} face detector to extract face crops. The data preprocessing produces image segments $\{s^{i}_{1},s^{i}_{2},...,s^{i}_{n}\}$ and the corresponding audio segments $\{a^{i}_{1},a^{i}_{2},...,a^{i}_{n}\}$, where $n$ denotes segment count for an input video $v^{i}$. The classifier consists of a bi-stream network, where each image segment $s^{i}_{t}$ ($t=1...n$) is passed through a video stream $S_{v}$, and the corresponding audio segment $a^{i}_{t}$ is passed through an audio stream $S_{a}$. The network is trained using a combination of contrastive loss ($L_1$) and cross-entropy loss ($L_2$ and $L_3$). The contrastive loss is meant to ensure that the video and audio streams are closer for real videos, and farther for fake videos. The cross-entropy loss is meant to ensure learning of robust video and audio feature representations. The overall loss $L$ is a weighted sum of the three losses, $L_1$, $L_2$ and $L_3$~\cite{Chugh2020Not}:

\begin{equation}
L_1=\frac{1}{N}\sum^N_{i=1}(y^i)(d^i_t)^2+(1-y^i)max(margin-d^i_t,0)^2
\end{equation}

\begin{equation}
d^i_t=||f_v-f_a||_2
\end{equation}

\begin{equation}
L_2=-\frac{1}{N}\sum^N_{i = 1}y^ilog\hat{y}^i_v+(1-y^i)log(1-\hat{y}^i_v)
\end{equation}

\begin{equation}
L_3=-\frac{1}{N}\sum^N_{i = 1}y^ilog\hat{y}^i_a+(1-y^i)log(1-\hat{y}^i_a)
\end{equation}

\begin{equation}
L=\lambda_1L_1+\lambda_2L_2+\lambda_3L_3
\end{equation}
where $\lambda_1$, $\lambda_2$ and $\lambda_3$ all equal to 1, $margin$ is a hyper-parameter and $f_v$ and $f_a$ are feature representations of the video and audio streams respectively.

This work extends~\cite{Chugh2020Not} with the introduction of novel visual representations of physiological signals into the network. The proposed visual representations are either used to augment the original face crops or the relationship between the face crops and the proposed visual representations is learned from data through a novel Graph Convolutional Network based architecture. The next subsections detail this further. An overview of the proposed method is illustrated in Figure~\ref{fig:graph_conv_arch}.

\subsection{Generation of Physiological Maps}
For the generation of visual representations of physiological signals we utilize recent advancements in camera-based physiological measurement. In particular, we leverage a novel Multi-Task Temporal Shift Convolutional Attention Network (MTTS-CAN)~\cite{Liu2020Multi} that enables real-time cardiovascular (\ie heart rate) and respiratory measurements (\ie respiration rate), that is,  given an RGB video of a frontal face (\ie face crop), the network estimates the waveform of the two physiological signals.

\begin{figure*}[ht]
\centering 
\includegraphics[width=\textwidth]{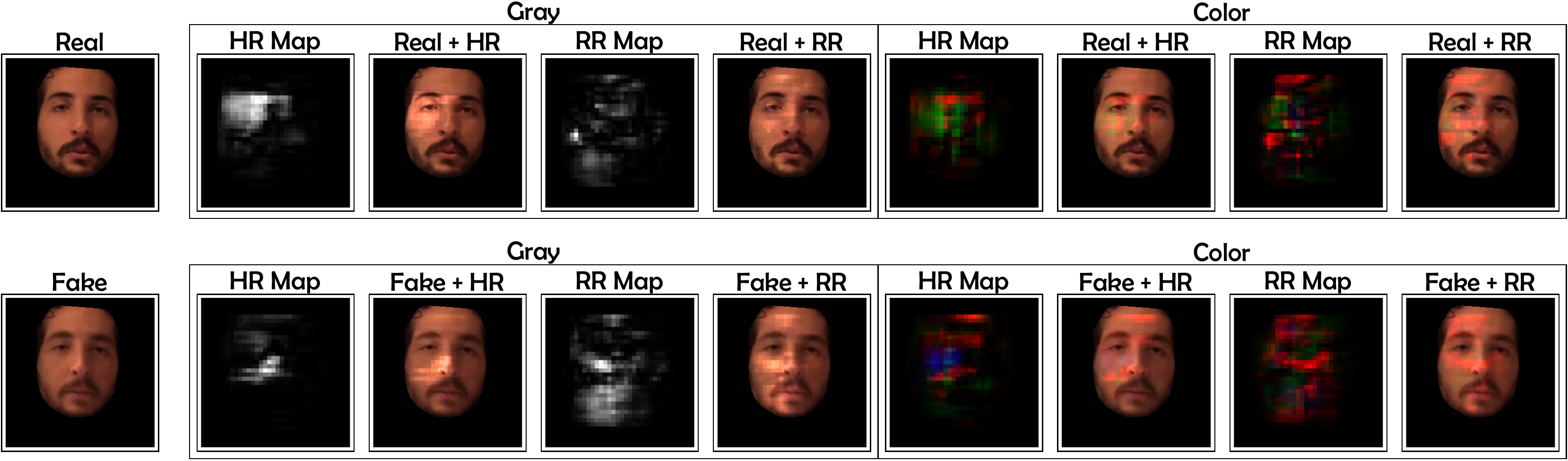}
\caption{Example of data augmentation. The real (top row) and the corresponding fake (bottom row) aligned face crops are augmented with the generated gray and color HR (\ie heart rate) and RR (\ie respiration rate) physiological maps.}
\label{fig:phys_maps_aug}
\end{figure*}

One of the contributions of this work is a method for generation of visual representations (\ie physiological maps) of physiological signals based on the estimated signal waveform. Visual representations of the signal waveform is important since a single float number estimate for the physiological signal for each face crop is not useful data representation. Therefore, we propose a novel method for generation of visual representations that can be used to learn relationships between the video and physiological signals and provide a significant improvement in the downstream task of fake video detection. The process of physiological maps generation is divided into several steps: 1) the face is detected and aligned with OpenFace~\cite{Baltrusaitis2018OpenFace} and most of the background is removed, 2) the aligned face crops are passed through MTTS-CAN and the two physiological signals (\ie heart rate and respiration rate) are estimated and kept as reference, 3) a square occlusion patch is defined and used to mask-out part of all aligned face crops and the occluded face crops are passed through MTTS-CAN to estimate the two physiological signals, 4) the difference between the estimated reference signals without occlusion and the estimated signals with occlusion is considered as the relative contribution (importance) of the masked-out region for the accurate estimation of the two physiological signals, and 5) the occlusion patch is moved and steps 3 and 4 are repeated. Additionally, the occlusion patch in step 3 is either applied to all color channels or separately for each color channel. This results in two sets of physiological maps -- \textit{gray} (a single difference value per patch) and \textit{color} (three difference values per patch for each color channel). The overall result of this process is the generation of visual representations that encode the relative contribution of all masked-out regions for the accurate estimation of the two physiological signals. Figure~\ref{fig:phys_maps_gen} offers a visualization of the steps involved in the generation of the proposed physiological maps.

\subsection{Physiological Maps Augmentation}
Given the visual representations of the physiological signals, one approach to utilize this additional information is to augment the aligned face crops. In particular, this work employs simple multiplication of the aligned face crops and the corresponding physiological maps as a data augmentation strategy. This process results in an augmented dataset that incorporates the information for the two physiological signals. An example of data augmentation is provided in Figure~\ref{fig:phys_maps_aug}.

\subsection{Cross-Modal Learning}
Given the physiological maps, another approach to utilize this additional information is to learn the relationship between the aligned face crops and physiological maps from data. This work proposes a multimodal fusion of the aligned face crops and physiological maps with a graph-based model. For a given video segment, each aligned face crop and the corresponding physiological map are used as nodes. The node features for both the crops and maps are obtained using ResNet-18~\cite{He2016Deep} (with the pre-trained weights on ImageNet~\cite{Deng2009Imagenet}). Inside the graph, each of the maps is connected to all crops with an edge with weight given by the cosine similarity between the node features. This representation enables the model to learn the weights of each map/crop. Given this graph representation, this work uses a Graph Convolutional Network (GCN)~\cite{kipf2017Semi} to learn the node features. Then the learned node features are concatenated with the features generated by the video stream of the bi-stream classifier network as shown in Figure~\ref{fig:graph_conv_arch}.

The graph-based multimodal feature fusion consists of $N$ nodes, $X=\{x_1, x_2,.., x_N\}$, corresponding to each physiological map and aligned face crop across $T$ time steps. Each of the map nodes is connected to all aligned face crop nodes. With this graph representation, a Graph Convolutional Network performs message passing based updates in the graph to update the features for each node. Formally, given a node feature matrix of dimension $N\times d$, there exists an edge between all the pairs of nodes of the form $<$physiological map node : aligned face crop node$>$; hence there are $T^2$ edges in the graph representation weighted by the similarity between the nodes they connect. The similarity criterion (\ie edge weight) between two nodes $x_1$ and $x_2$ is defined by the cosine similarity between the node features:

\begin{equation}
EdgeWeight(x_1,x_2)=\frac{\sum^d_{i=1}{x_1}_i{x_2}_i}{ \sqrt{\sum^d_{i=1}({x_1}_i)^2}\sqrt{\sum^d_{i=1}({x_2}_i)^2}}
\end{equation}

\section{Experiments}
\label{sec:experiments}
This section describes the datasets used and the general setup of the experiments.

\subsection{Datasets}
The deepfake detection challenge (DFDC) dataset was released gradually with the preview dataset~\cite{Dolhansky2019Preview}, comprising of 5214 videos and the complete dataset~\cite{Dolhansky2020DFDC} with 119146 videos. The details of the video manipulations were not disclosed in order to represent the real adversarial space of facial manipulation. The manipulations can be present in either the audio or video or both of the channels. In order to bring out a fair comparison, this work uses 18000 videos\footnote{Identical to those used in \cite{Chugh2020Not} and \cite{Mittal2020Emotions}} of DFDC in all experiments. The videos are of $\approx10$ seconds duration each with a frame rate of 30 frames per second.

The Deepfake-TIMIT~\cite{Korshunov2018DeepFakes} dataset contains videos of 16 similar looking pairs of people, which are manually selected from the publicly available VIDTIMIT\footnote{https://conradsanderson.id.au/vidtimit} dataset and manipulated using an open-source Generative Adversarial Network based approach. There are two different models for generating fake videos, one low quality (LQ), with $64\times64$ input/output size, and the other high quality (HQ), with $128\times128$ input/output size. Each of the 32 subjects has 10 videos, resulting in a total of 640 fake videos in the dataset. The videos are of $\approx4$ seconds duration each with a frame rate of 25 frames per second. In this dataset the audio channel is not manipulated.

\subsection{Implementation}
In all experiments the hyper-parameters are kept the same as in~\cite{Chugh2020Not} that were arrived at by a series of ablation studies. The segment duration $D$ is 1 second and the margin hyper-parameter for the contrastive loss is set to 0.99. The graph is composed of $30*30$ edges (with $T=30$ time steps for DFDC) and $25*25$ edges (with $T=25$ time steps for Deepfake-TIMIT). The node features size is $d=512$ and the Graph Convolutional Network layer size is $(512, 16)$. The physiological map generation is performed on $36\times36$ down-sampled versions of the aligned face crops since this is the input size expected by the MTTS-CAN model. The occlusion patch size is $9\times9$ resulting in $36\times36\times3$ (color) or $36\times36\times1$ (gray) dimensional physiological maps. The models are trained with a batch size of 8 using the Adam~\cite{Kingma2014Adam} optimizer with default parameters ($\alpha=0.001$, $\beta_{1}=0.9$, $\beta_{2}=0.999$, and $\epsilon=10^{-8}$) and with a learning rate of 0.001. All models are implemented in PyTorch~\cite{Paszke2017Automatic}.

\subsection{Evaluation}
Similarly to~\cite{Chugh2020Not}, during model evaluation, the video segments $\{s^{i}_{1},s^{i}_{2},...,s^{i}_{n}\}$ and corresponding
audio segments $\{a^{i}_{1},a^{i}_{2},...,a^{i}_{n}\}$ of a test video are passed through $S_{v}$ and $S_{a}$. Then for each segment, a dissimilarity score $d^{i}_{t}=||f_{v}-f_{a}||_{2}$ is accumulated, where $f_{v}$ and $f_{a}$ are feature representations of the video and audio streams, respectively. Finally, the Modality Dissonance Score (MDS) is computed as the average of the accumulated dissimilarity scores:

\begin{equation}
MDS_{i}=\frac{1}{n}\sum^{n}_{t=1}{d^{i}_{t}}
\end{equation}

In order to classify a test video as either real or fake, the $MDS_{i}$ is compared with a threshold $\tau$ using $I\{MDS_{i} < \tau\}$, where $I\{.\}$ denotes the logical indicator function and $\tau$ is determined using the train set. Then the Area Under the Curve (AUC) metric is used for evaluation, computed using video-wise real/fake classification. This evaluation strategy is consistent with the one in~\cite{Chugh2020Not} and ensures the fair comparison between the performance of the different methods.

\section{Results}
\label{sec:results}
This section reports that the proposed physiology-based fake video detection method achieves competitive performance on both the DFDC and Deepfake-TIMIT datasets. Table~\ref{tab:results}, starting from second row, reports the classification results on the test set of the DFDC and DF-TIMIT datasets using the proposed data augmentation strategy. The results from MDS models using three of the proposed physiological maps for data augmentation are in par with the MDS method (without augmentation). The second and third column present the detection results on the test set of the Deepfake-TIMIT HQ and LQ dataset, respectively. Here the results from all MDS models using physiological maps for data augmentation are better than the original MDS method (\ie 96.8\% for HQ and 97.9\% for LQ).

\begin{table}[ht!]
\caption{Comparison of the video-level classification performance using the proposed physiological maps for data augmentation. The MDS models are evaluated on the DFDC and Deepfake-TIMIT HQ/LQ datasets with \%AUC metric.}
\label{tab:results}
\begin{center}
\begin{tabular}{c | c c c}
\textbf{Maps/Datasets} & DFDC & DF-TIMIT (HQ) & DF-TIMIT (LQ)\\
\hline
Not Augmented~\cite{Chugh2020Not} & 91.5 & 96.8 & 97.9\\\\
HR Gray Augmented & 91.4 & 98.1 & 99.7\\
HR Color Augmented & \textbf{91.6} & 99.2 & \textbf{1.0}\\
RR Gray Augmented & 91.5 & \textbf{1.0} & 99.2\\
RR Color Augmented & 89.8 & 99.4 & \textbf{1.0}
\end{tabular}
\end{center}
\end{table}

Comparison of the proposed learning-based fusion method and previous methods in terms of classification performance is provided in Table~\ref{tab:results_all}. The table reports the performance of the best performing fusion method with HR color physiological maps. Using the HR (\ie heart rate) color maps (the best performing maps during data augmentation) for learning the fusion between physiological maps and aligned face crops with the Graph Convolutional Network yields performance of 93.1\% which is better than the MDS method with 1.6\%. By extension, this suggests that the proposed method outperforms the other visual-only and audio-visual approaches discussed in~\cite{Chugh2020Not}.

We note that the results from Graph Convolutional Network based model on Deepfake-TIMIT dataset are slightly worse than but comparable to the proposed approach using physiological maps for augmentation as given in Table~\ref{tab:results}. Since the Deepfake-TIMIT dataset is a small dataset (only 1281 test video samples), the results are highly oscillating and a few misclassifications result in change of evaluation metric scores. Augmentation of the video frames with the generated physiological maps performs better on these datasets. Overall, methods involving fusion with HR Color maps either directly augmented or the relationships learned through Graph Convolutional Network, outperform previous methods as seen in Table~\ref{tab:results_all}.

\begin{table}[ht!]
\caption{Comparison of the video-level classification performance using the proposed physiological maps for GCN-based learning with other methods on the DFDC and Deepfake-TIMIT HQ/LQ datasets using \%AUC metric.}
\label{tab:results_all}
\begin{center}
\begin{tabular}{c | c c c}
\textbf{Methods/Datasets} & DFDC & DF-TIMIT (HQ) & DF-TIMIT (LQ)\\
\hline
Capsule~\cite{Nguyen2019Capsule} & 53.3 & 74.4 & 78.4\\
Multi-task~\cite{Nguyen2019Multi} & 53.6 & 55.3 & 62.2\\
HeadPose~\cite{Yang2019Exposing} & 55.9 & 53.2 & 55.1\\
Two-stream~\cite{Zhou2018Two} & 61.4 & 73.5 & 83.5\\
VA-LogReg~\cite{Matern2019Exploiting} & 66.2 & 77.3 & 77.0\\
Meso4~\cite{Afchar2018MesoNet} & 75.3 & 68.4 & 87.8\\
Xception-c23~\cite{Roessler2019FaceForensics} & 72.2 & 94.4 & 95.9\\
DSP-FWA~\cite{Li2019Exposing} & 75.5 & \textbf{99.7} & \textbf{99.9}\\
Siamese~\cite{Mittal2020Emotions} & 84.4 & 94.9 & 96.3\\
MDS~\cite{Chugh2020Not} & 91.5 & 96.8 & 97.9\\
\hline\hline
\textbf{Ours} &  \textbf{93.1} & 96.2 & 97.8
\end{tabular}
\end{center}
\end{table}

\section{Discussion}
\label{sec:discussion}
Along with leading to an improved performance on AUC metric, our graph-based fusion framework is an interpretable model, that is, the edges in the graph (and hence the frames) contributing most to the given model decision can be identified based on the edge weights. The framework can be helpful in making insightful progress on the misclassified samples by providing frame-based information from the graph edge weights joining two modalities. This interpretability aspect can be very useful in real-world setting where it becomes important to work out the causes of misclassifications.

The current work investigates the fusion/augmentation of only one physiological measurement with the video stream. Given that both physiological signals might provide useful and complementary information for the task of fake video detection, a direction for future work includes fusion of both types of physiological masks and video data at the same time. 

MTTS-CAN used for estimation of the heart and respiration rate has been shown to perform well for frontal faces with limited movement. However, the data in the DFDC dataset is unconstrained and includes lots of head movement, lighting conditions, and backgrounds. We partially address this problem with the face alignment step to provide higher quality data for MTTS-CAN. We expect that the fake video classification performance will be even higher in case more advanced and accurate methods are used for alignment and estimation of physiological signals.

Finally we note that in~\cite{Chugh2020Not} one of the arguments for the increased performance of the MDS model is due to the relaxed crop around the face region. This suggests that the contribution of the proposed physiological maps is bigger than the 1.6\% because the visual inputs in this work are aligned faces (with removed background). Indeed, re-training the original MDS model with aligned faces yields a performance of 88.6\% on the DFDC dataset which implies that the relative contribution of the HR color maps is an increase of 4.5\%.

\section{Conclusion}
\label{sec:conclusion}
We propose a novel method for generation of visual representations of physiological signals extracted from facial videos. Additionally, we propose two strategies for utilizing those representations for the problem fake video detection, either by augmenting the video modality with information from the physiology or by novelly learning the fusion of those two modalities with a Graph Convolutional Network. Experiments show that this novel physiological maps help to achieve competitive performance. Future work will focus on the simultaneous fusion of both types of physiological maps and video stream and investigation on different similarity measurements used in the Graph Convolutional Network.



\bibliographystyle{IEEEtran}
\bibliography{paper}

\begin{thebibliography}{10}
\providecommand{\url}[1]{#1}
\csname url@samestyle\endcsname
\providecommand{\newblock}{\relax}
\providecommand{\bibinfo}[2]{#2}
\providecommand{\BIBentrySTDinterwordspacing}{\spaceskip=0pt\relax}
\providecommand{\BIBentryALTinterwordstretchfactor}{4}
\providecommand{\BIBentryALTinterwordspacing}{\spaceskip=\fontdimen2\font plus
\BIBentryALTinterwordstretchfactor\fontdimen3\font minus
  \fontdimen4\font\relax}
\providecommand{\BIBforeignlanguage}[2]{{%
\expandafter\ifx\csname l@#1\endcsname\relax
\typeout{** WARNING: IEEEtran.bst: No hyphenation pattern has been}%
\typeout{** loaded for the language `#1'. Using the pattern for}%
\typeout{** the default language instead.}%
\else
\language=\csname l@#1\endcsname
\fi
#2}}
\providecommand{\BIBdecl}{\relax}
\BIBdecl

\bibitem{Dolhansky2020DFDC}
B.~Dolhansky, J.~Bitton, B.~Pflaum, J.~Lu, R.~Howes, M.~Wang, and C.~C. Ferrer,
  ``The deepfake detection challenge dataset,'' 2020.

\bibitem{Korshunov2018DeepFakes}
P.~Korshunov and S.~Marcel, ``Deepfakes: a new threat to face recognition?
  assessment and detection,'' 2018.

\bibitem{Roessler2019FaceForensics}
A.~R\"ossler, D.~Cozzolino, L.~Verdoliva, C.~Riess, J.~Thies, and
  M.~Nie{\ss}ner, ``Face{F}orensics++: Learning to detect manipulated facial
  images,'' in \emph{Proceedings of the International Conference on Computer
  Vision}, 2019.

\bibitem{Verdoliva2019Multimedia}
L.~Verdoliva and P.~Bestagini, ``Multimedia forensics,'' in \emph{Proceedings
  of the ACM International Conference on Multimedia}, 2019, pp. 2701--2702.

\bibitem{Yang2019Exposing}
X.~Yang, Y.~Li, and S.~Lyu, ``Exposing deep fakes using inconsistent head
  poses,'' in \emph{Proceedings of the IEEE International Conference on
  Acoustics, Speech and Signal Processing}, 2019, pp. 8261--8265.

\bibitem{Zheng2019Survey}
L.~Zheng, Y.~Zhang, and V.~L. Thing, ``A survey on image tampering and its
  detection in real-world photos,'' \emph{Journal of Visual Communication and
  Image Representation}, vol.~58, pp. 380--399, 2019.

\bibitem{Mirsky2021Creation}
Y.~Mirsky and W.~Lee, ``The creation and detection of deepfakes: A survey,''
  \emph{ACM Computing Surveys}, 2021.

\bibitem{Mo2018Fake}
H.~Mo, B.~Chen, and W.~Luo, ``Fake faces identification via convolutional
  neural network,'' in \emph{Proceedings of the ACM Workshop on Information
  Hiding and Multimedia Security}, 2018, pp. 43--47.

\bibitem{Durall2020Unmasking}
R.~Durall, M.~Keuper, F.-J. Pfreundt, and J.~Keuper, ``Unmasking deepfakes with
  simple features,'' 2020.

\bibitem{Li2019Exposing}
Y.~Li and S.~Lyu, ``Exposing deepfake videos by detecting face warping
  artifacts,'' in \emph{Proceedings of the IEEE Conference on Computer Vision
  and Pattern Recognition Workshops}, 2019.

\bibitem{Straub2019Using}
J.~Straub, ``Using subject face brightness assessment to detect `deep fakes',''
  in \emph{Proceedings of the Real-Time Image Processing and Deep Learning},
  2019.

\bibitem{Agarwal2019Protecting}
S.~Agarwal, H.~Farid, Y.~Gu, M.~He, K.~Nagano, and H.~Li, ``Protecting {World}
  {Leaders} {Against} {Deep} {Fakes},'' in \emph{Proceedings of the {IEEE}
  {Conference} on {Computer} {Vision} and {Pattern} {Recognition} {Workshops}},
  2019.

\bibitem{Ciftci2020FakeCatcher}
U.~A. Ciftci, I.~Demir, and L.~Yin, ``Fakecatcher: Detection of synthetic
  portrait videos using biological signals,'' \emph{IEEE Transactions on
  Pattern Analysis and Machine Intelligence}, 2020.

\bibitem{Korshunov2018Speaker}
P.~Korshunov and S.~Marcel, ``Speaker inconsistency detection in tampered
  video,'' in \emph{Proceedings of the European Signal Processing Conference},
  2018, pp. 2375--2379.

\bibitem{Guera2018Deepfake}
D.~Guera and E.~J. Delp, ``Deepfake video detection using recurrent neural
  networks,'' in \emph{Proceedings of the IEEE International Conference on
  Advanced Video and Signal Based Surveillance}, 2018, pp. 1--6.

\bibitem{Afchar2018MesoNet}
D.~Afchar, V.~Nozick, J.~Yamagishi, and I.~Echizen, ``Mesonet: a compact facial
  video forgery detection network,'' in \emph{Proceedings of the IEEE Workshop
  on Information Forensics and Security}, 2018, pp. 1--7.

\bibitem{Ding2019Swapped}
X.~Ding, Z.~Raziei, E.~C. Larson, E.~V. Olinick, P.~Krueger, and M.~Hahsler,
  ``Swapped face detection using deep learning and subjective assessment,''
  2019.

\bibitem{Tariq2018Detecting}
S.~Tariq, S.~Lee, H.~Kim, Y.~Shin, and S.~S. Woo, ``Detecting both machine and
  human created fake face images in the wild,'' in \emph{Proceedings of the
  International Workshop on Multimedia Privacy and Security}, 2018, pp. 81--87.

\bibitem{Wang2020FakeSpotter}
R.~Wang, F.~Juefei-Xu, L.~Ma, X.~Xie, Y.~Huang, J.~Wang, and Y.~Liu,
  ``Fakespotter: A simple yet robust baseline for spotting ai-synthesized fake
  faces,'' 2020.

\bibitem{Khalid2020FakeDect}
H.~Khalid and S.~S. Woo, ``Oc-fakedect: Classifying deepfakes using one-class
  variational autoencoder,'' in \emph{Proceedings of the IEEE/CVF Conference on
  Computer Vision and Pattern Recognition Workshops}, 2020, pp. 2794--2803.

\bibitem{Liu2020Multi}
X.~Liu, J.~Fromm, S.~N. Patel, and D.~McDuff, ``Multi-task temporal shift
  attention networks for on-device contactless vitals measurement,'' 2020.

\bibitem{Chugh2020Not}
K.~Chugh, P.~Gupta, A.~Dhall, and R.~Subramanian, ``Not made for each other-
  audio-visual dissonance-based deepfake detection and localization,'' in
  \emph{Proceedings of the ACM International Conference on Multimedia}, 2020,
  pp. 439--447.

\bibitem{Zhang2017S3FD}
S.~Zhang, X.~Zhu, Z.~Lei, H.~Shi, X.~Wang, and S.~Li, ``S3fd: Single shot
  scale-invariant face detector,'' in \emph{Proceedings of the IEEE
  International Conference on Computer Vision}, 2017, pp. 192--201.

\bibitem{Baltrusaitis2018OpenFace}
T.~Baltrusaitis, A.~Zadeh, Y.~C. Lim, and L.-P. Morency, ``Openface 2.0: Facial
  behavior analysis toolkit,'' in \emph{Proceedings of the IEEE International
  Conference on Automatic Face and Gesture Recognition}, 2018, pp. 59--66.

\bibitem{He2016Deep}
K.~He, X.~Zhang, S.~Ren, and J.~Sun, ``Deep residual learning for image
  recognition,'' in \emph{Proceedings of the IEEE Conference on Computer Vision
  and Pattern Recognition}, 2016, pp. 770--778.

\bibitem{Deng2009Imagenet}
J.~Deng, W.~Dong, R.~Socher, L.-J. Li, K.~Li, and L.~Fei-Fei, ``Imagenet: A
  large-scale hierarchical image database,'' in \emph{Proceedings of the IEEE
  Conference on Computer Vision and Pattern Recognition}, 2009, pp. 248--255.

\bibitem{kipf2017Semi}
T.~N. Kipf and M.~Welling, ``Semi-supervised classification with graph
  convolutional networks,'' 2017.

\bibitem{Dolhansky2019Preview}
B.~Dolhansky, R.~Howes, B.~Pflaum, N.~Baram, and C.~C. Ferrer, ``The deepfake
  detection challenge (dfdc) preview dataset,'' 2019.

\bibitem{Mittal2020Emotions}
T.~Mittal, U.~Bhattacharya, R.~Chandra, A.~Bera, and D.~Manocha, ``Emotions
  don't lie: An audio-visual deepfake detection method using affective cues,''
  in \emph{Proceedings of the ACM International Conference on Multimedia},
  2020, pp. 2823--2832.

\bibitem{Kingma2014Adam}
D.~P. Kingma and J.~Ba, ``Adam: a method for stochastic optimization,'' 2014.

\bibitem{Paszke2017Automatic}
A.~Paszke, S.~Gross, S.~Chintala, G.~Chanan, E.~Yang, Z.~DeVito, Z.~Lin,
  A.~Desmaison, L.~Antiga, and A.~Lerer, ``Automatic differentiation in
  {PyTorch},'' in \emph{Proceedings of the NeurIPS Autodiff Workshop}, 2017.

\bibitem{Nguyen2019Capsule}
H.~H. Nguyen, J.~Yamagishi, and I.~Echizen, ``Capsule-forensics: Using capsule
  networks to detect forged images and videos,'' in \emph{Proceedings of the
  IEEE International Conference on Acoustics, Speech and Signal Processing},
  2019, pp. 2307--2311.

\bibitem{Nguyen2019Multi}
H.~H. Nguyen, F.~Fang, J.~Yamagishi, and I.~Echizen, ``Multi-task learning for
  detecting and segmenting manipulated facial images and videos,'' in
  \emph{Proceedings of the IEEE International Conference on Biometrics Theory,
  Applications and Systems}, 2019, pp. 1--8.

\bibitem{Zhou2018Two}
P.~Zhou, X.~Han, V.~I. Morariu, and L.~S. Davis, ``Two-stream neural networks
  for tampered face detection,'' 2018.

\bibitem{Matern2019Exploiting}
F.~Matern, C.~Riess, and M.~Stamminger, ``Exploiting visual artifacts to expose
  deepfakes and face manipulations,'' in \emph{Proceedings of the IEEE Winter
  Applications of Computer Vision Workshops}, 2019, pp. 83--92.

\end{thebibliography}

\end{document}